\documentclass{article}
\usepackage{spconf,amsmath,graphicx}
\usepackage{amsfonts,multirow,tabularx}
\usepackage{url}
\usepackage{caption}
\usepackage{cite}
\usepackage{fancyhdr}

\def\cpar{\hss\egroup\line\bgroup\hss}

\title{Fake Generated Painting Detection via Frequency Analysis }

\name{Yong Bai$^{\star}$ \qquad  Yuanfang Guo$^{\star}$ \qquad Jinjie Wei$^{\star}$ 
 \qquad Lin Lu$^{\star}$ \qquad Rui Wang$^{\dagger}$ $^{\ddagger}$ \qquad Yunhong Wang$^{\star}$
 \thanks{This work was supported in part by the National Natural Science Foundation of China under Grant 61802391 and Grant 61421003, in part by the Fundamental Research Funds for Central Universities.}
 \thanks{The corresponding author is Yuanfang Guo.}
 \thanks{(Email: andyguo@buaa.edu.cn)}
} 
\address{$^{\star}$ School of Computer Science and Engineering, \\ 
Beihang University, Beijing, China \\
    $^{\dagger}$ State Key Laboratory of Information Security, \\ 
    Institute of Information Engineering, Chinese Academy of Sciences, Beijing, China \\
    $^{\ddagger}$ School of Cyber Security, University of Chinese Academy of Sciences, Beijing, China
    }

\begin{document}
\topmargin=0mm
%\ninept
\maketitle

\thispagestyle{fancy}
\fancyhead{}
\lhead{}
\lfoot{ \copyright 2020 IEEE. Personal use of this material is permitted. Permission from IEEE must be obtained for all other uses, in any current or future media, including reprinting/republishing this material for advertising or promotional purposes, creating new collective works, for resale or redistribution to servers or lists, or reuse of any copyrighted component of this work in other works.}
\cfoot{}
\rfoot{}

\begin{abstract}
With the development of deep neural networks, digital fake paintings can be generated by various style transfer 
algorithms. 
To detect the fake generated paintings, we analyze the fake generated and real paintings in Fourier frequency domain 
and observe statistical differences and artifacts. Based on our observations, we propose Fake Generated Painting Detection 
via Frequency Analysis (FGPD-FA) by extracting three types of features in frequency domain. 
Besides, we also propose a digital fake painting detection database for assessing the proposed method. 
Experimental results demonstrate the excellence of the proposed method in different testing conditions.

\end{abstract}
\section{INTRODUCTION}
\label{sec:intro}

Painting has been a major type of artworks in the past few centuries. Many artists, such as Van Gogh, Monet, etc., 
have contributed numerous paintings with their unique styles. Recently, deep neural networks are developed to transfer the styles 
of the artists/paintings to other natural images and thus generate realistic digital paintings. This generation can mainly be achieved via two mechanisms. 
On one hand, 
\cite {DBLP:conf/cvpr/GatysEB16,DBLP:conf/eccv/JohnsonAF16,DBLP:conf/icml/UlyanovLVL16,DBLP:journals/corr/ChenS16f,DBLP:conf/cvpr/YaoRX0LW19}
propose that content and style can be reconstructed based on the image features extracted from convolutional neural networks (CNN). On the other hand, the emergence of generative adversarial networks (GAN) 
provides another approach to perform style transfer  \cite {DBLP:conf/iccv/ZhuPIE17,DBLP:conf/icml/KimCKLK17,DBLP:journals/tip/ChenXYST19}.

\begin{figure}
    \centering
    \begin{minipage}[b]{0.24\linewidth}
      \centerline{\includegraphics[width=\linewidth]{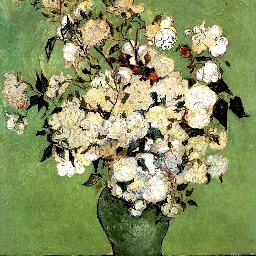}}
      \centerline{(a)}
    \end{minipage}
    \begin{minipage}[b]{0.24\linewidth}
      \centerline{\includegraphics[width=\linewidth]{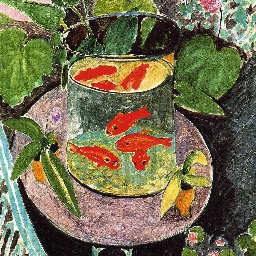}}
      \centerline{(b)}
    \end{minipage}
  %  \vspace{2.0cm}
    % \medskip
    \begin{minipage}[b]{0.24\linewidth}
      \centerline{\includegraphics[width=\linewidth]{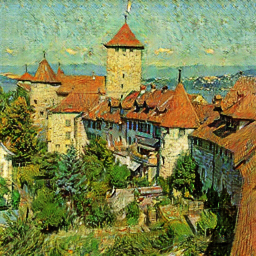}}
      \centerline{(c)}
    \end{minipage}
    \begin{minipage}[b]{0.24\linewidth}
      \centerline{\includegraphics[width=\linewidth]{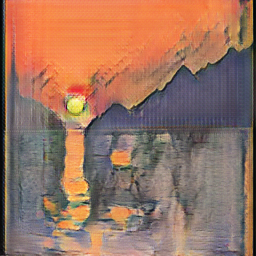}}
      \centerline{(d)}
    \end{minipage}
  %  \vspace{2.0cm}
  
    % \medskip
  
  \caption{Examples of real and fake generated paintings:
  (a) Real, Van Gogh's style; (b) Real, Matisse's style; (c) Fake, Van Gogh's style, generated by CycleGAN \cite{DBLP:conf/iccv/ZhuPIE17};
  (d) Fake, Matisse's style, generated by CycleGAN \cite{DBLP:conf/iccv/ZhuPIE17}.}
  \label{fig:res}
\end{figure}

As shown in Fig.\,\ref{fig:res}, the generated paintings are also decent and can hardly be distinguished from 
the artists' work by ordinary people, i.e., people who lack the necessary expertise. 
Unfortunately, these generated paintings may be utilized by some malicious people to cheat the customers.
 Meanwhile, the artists' rights may also be violated. Therefore, it is necessary to identify whether a painting is generated (forged).

Unfortunately, traditional image forgery detection techniques are designed for traditional forgeries including copy-move, splicing and image retouching, which are different from style transfer in principle. 
Recently, several approaches have been proposed to detect the emerging forgeries, such as colorized images \cite {DBLP:journals/tifs/GuoCZW18}, 
GAN generated faces \cite {DBLP:journals/corr/abs-1808-07276,DBLP:journals/corr/abs-1912-13458} and computer graphics based generated images \cite {DBLP:journals/sensors/YaoHZWS18}. 
However, they may also be inappropriate, because they are designed to distinguish natural images, which are different from paintings, from fake images based on the corresponding forging mechanisms.
Currently, to the best of our knowledge, there exists no forgery detection method specifically designed for fake generated painting detection.

Considering that the existing style transfer techniques only possess constraints designed in spatial domain, the fake generated paintings may be more distinguishable 
in frequency domain. Therefore, we transform the real and fake generated paintings into Fourier domain to perform a frequency analysis. Based on our observation that the 
fake generated paintings reveal obvious inconsistencies in frequency domain compared to real paintings, we propose a forgery detection method, named Fake Generated 
Painting Detection via Frequency Analysis (FGPD-FA).

Our major contributions can be summarized as below:
1) We observe that there are differences between the real and fake generated paintings in frequency domain;
2) Based on the observed statistical differences between the spectrums of the real and fake generated paintings, we
extract the statistical and oriented gradient features for detection;
3) Based on the observed artifacts in the spectrum of fake generated paintings, we adopt a blob detection 
algorithm \cite{DBLP:journals/ijcv/Lowe04} to extract the blob feature for detection;
4) We build a painting database, named Digital Painting Forgery Detection (DPFD) Database, which contains 10310 real paintings from fifteen famous artists and 58292 fake paintings generated via
three state-of-the-art style transfer methods \cite {DBLP:conf/iccv/ZhuPIE17,DBLP:journals/tip/ChenXYST19,DBLP:conf/cvpr/YaoRX0LW19}.

\section{METHODOLOGY}
\label{sec:pagestyle}
\begin{figure}
  
  \begin{minipage}[b]{\linewidth}
    \centering
    \centerline{\includegraphics[width=\linewidth]{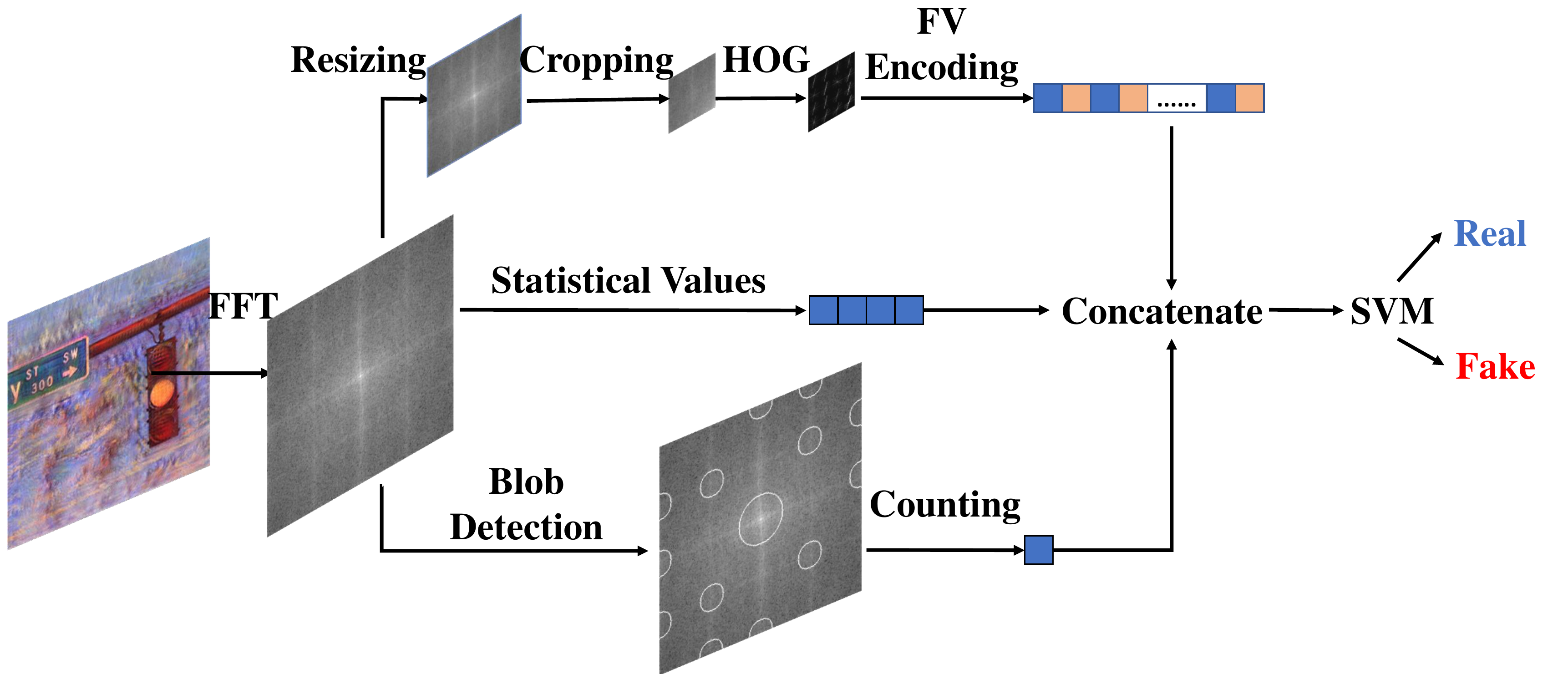}}
  %  \vspace{2.0cm}
    % \centerline{(a)}\medskip
  \end{minipage}
  
  \caption{Framework of the proposed FGPD-FA.}
  \label{fig:framework}
  \end{figure}

In this section, the proposed Fake Generated Painting Detection via Frequency Analysis (FGPD-FA), 
whose framework is shown in Fig.\,\ref{fig:framework}, will be introduced in details.

\subsection{Analysis and Observations}
\label{sec:observation}

\begin{figure}
  \centering
  \begin{minipage}[b]{0.24\linewidth}
    \centerline{\includegraphics[width=\linewidth]{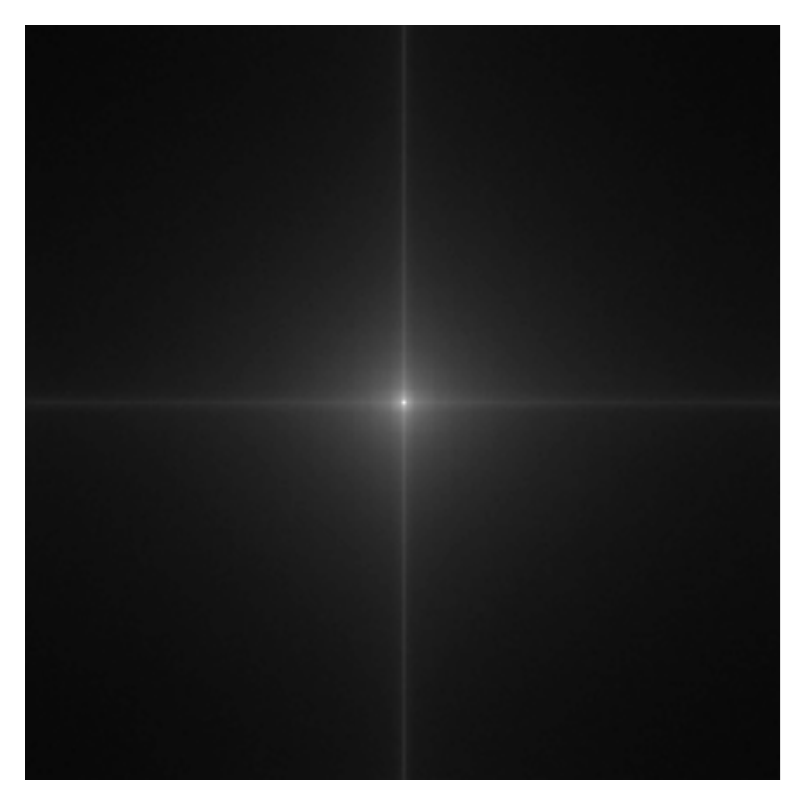}}
    \centerline{(a) Real}
  \end{minipage}
  \begin{minipage}[b]{0.24\linewidth}
    \centerline{\includegraphics[width=\linewidth]{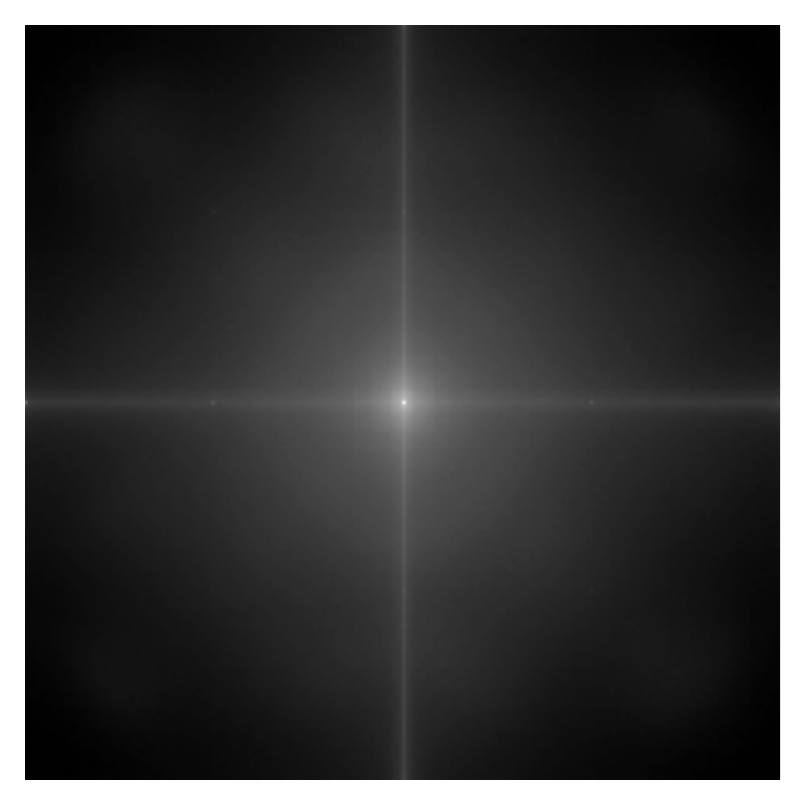}}
    \centerline{(b) AAMS}
  \end{minipage}
%  \vspace{2.0cm}
  % \medskip
  \begin{minipage}[b]{0.24\linewidth}
    \centerline{\includegraphics[width=\linewidth]{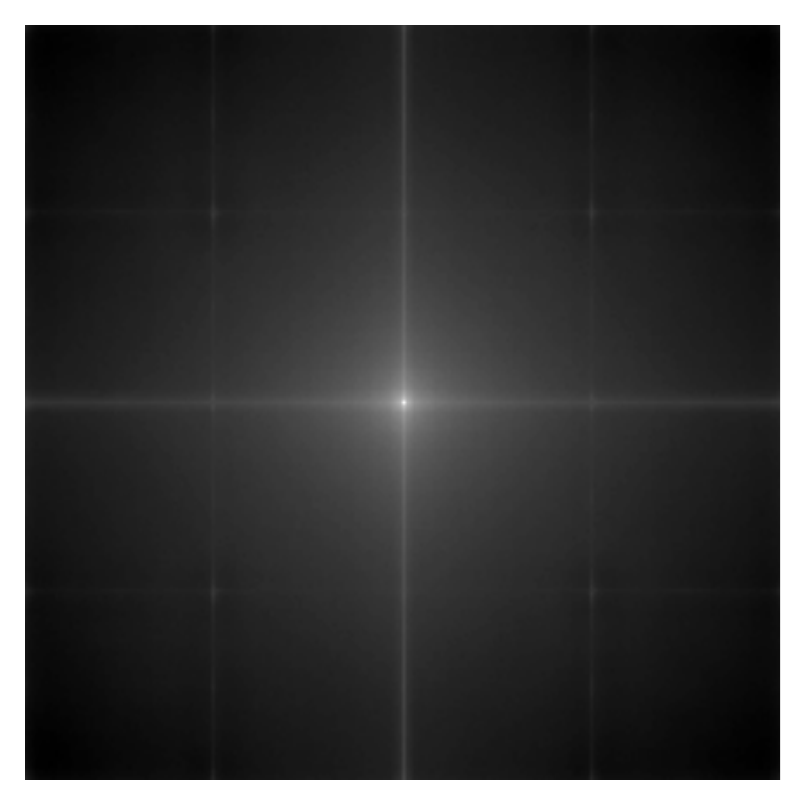}}
    \centerline{(c) CycleGAN}
  \end{minipage}
  \begin{minipage}[b]{0.24\linewidth}
    \centerline{\includegraphics[width=\linewidth]{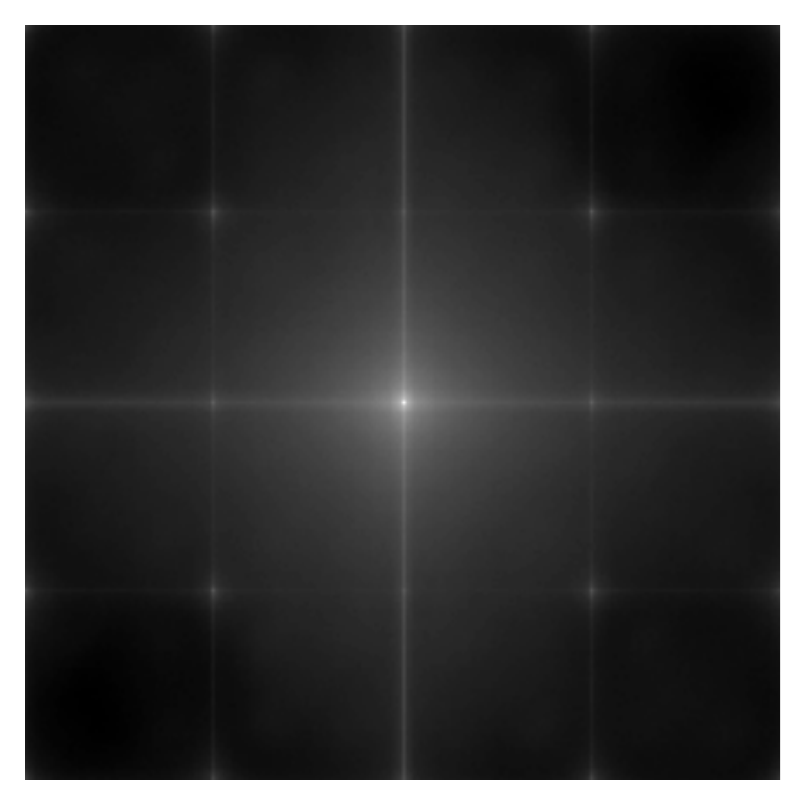}}
    \centerline{(d) GatedGAN}
  \end{minipage}
%  \vspace{2.0cm}

  % \medskip
  \caption{Averaged spectrum of the paintings.}
  \label{fig:spectrum}
\end{figure}

\begin{figure}
  % \begin{minipage}[b]{\linewidth}
    \centering
    \centerline{\includegraphics[width=0.7\linewidth]{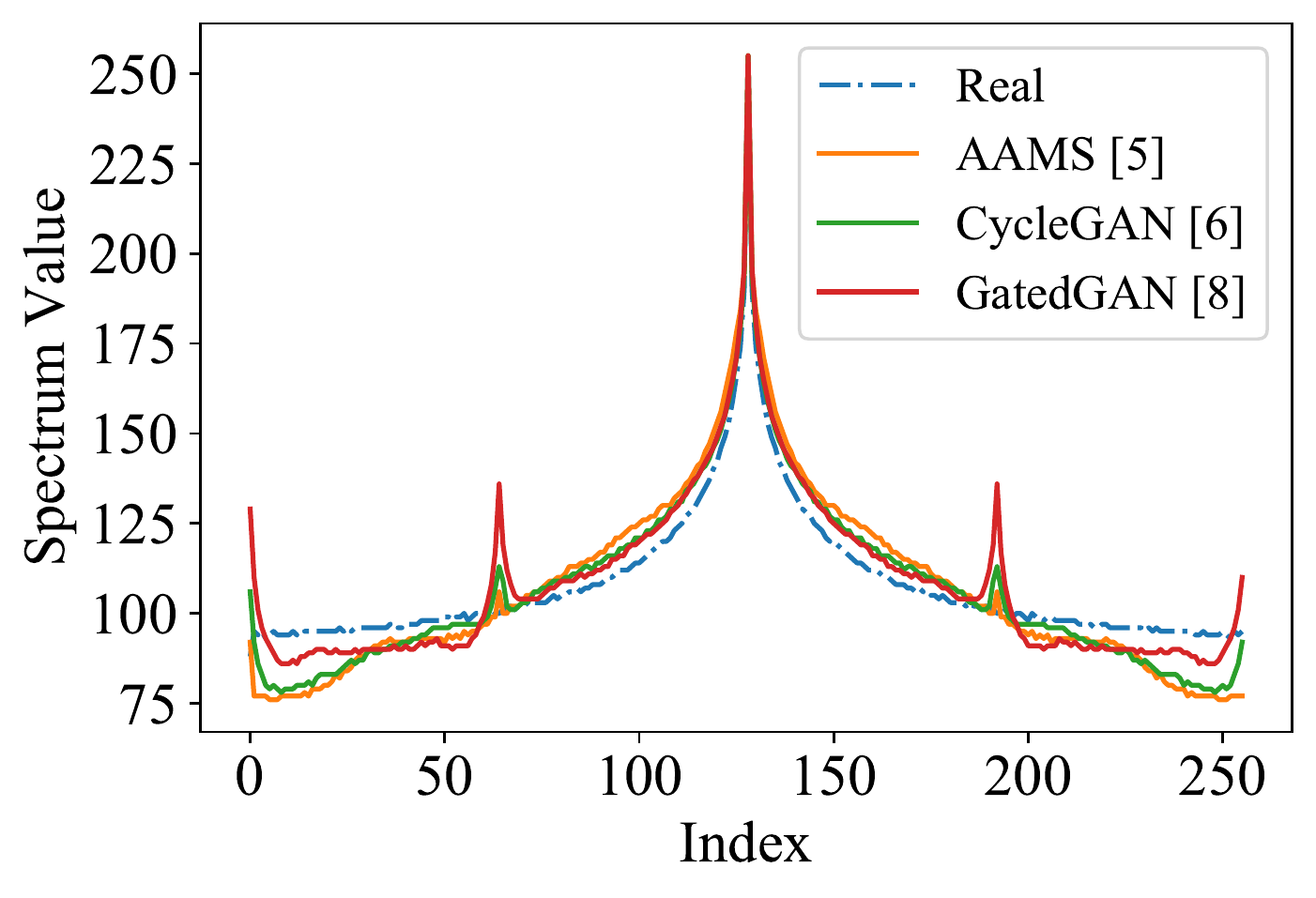}}
  %  \vspace{2.0cm}
    % \centerline{(a)}\medskip
  % \end{minipage}
  \caption{
    Values of the averaged spectrum's diagonal line.
  }
  \label{fig:stat}
\end{figure}

The differences between the fake generated and real paintings are not obvious, as Fig.\,\ref{fig:res} demonstrates. 
Since the state-of-the-art style transfer techniques do not constrain their outputs in frequency domain,
 a frequency analysis is performed to better reveal the differences between the fake generated and real paintings.
Naturally, Fourier transform is employed to transfer the paintings from spatial domain to frequency domain. 
Note that the colour space of the painting image will be transferred from RGB space to YUV space and only the 
Y component will be employed in the frequency analysis, for convenience. According to the experiments, 
two major observations can be obtained:

1) Statistical differences: The averaged distribution of the spectrum values are different between the fake generated and real paintings.
We randomly select 1000 spectrums of the real paintings and 3000 spectrums of the fake generated paintings, 
where each one third of the fake paintings are generated by a state-of-the-art forging method. 
Then, we show the averaged spectrum in Fig.\,\ref{fig:spectrum} and the values on the diagonal line of the averaged spectrum in Fig.\,\ref{fig:stat}.
As shown in Fig.\,\ref{fig:spectrum}, the difference of averaged spectrum values at middle frequencies between the fake generated and real paintings is obvious.
In Fig.\,\ref{fig:stat}, the x-axis represents 
the indexes of the diagonal line from the top-left corner to the bottom-right corner, while the y-axis represents the averaged spectrum values.
As can be observed, the spectrum values of the real paintings possess less variations and do not present spike-like artifacts compared to that of the fake generated paintings.

\begin{figure}
  \begin{minipage}[b]{\linewidth}
    \begin{minipage}[b]{0.6\linewidth}
      \centering
      \centerline{\includegraphics[width=\linewidth]{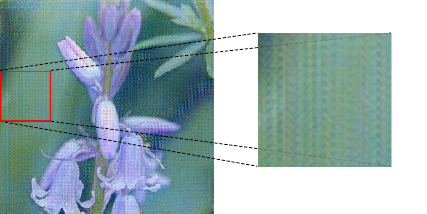}}
    %  \vspace{2.0cm}
      % \centerline{(a)}\medskip
    \end{minipage}
    \begin{minipage}[b]{0.3\linewidth}
      \centering
      \centerline{\includegraphics[width=\linewidth]{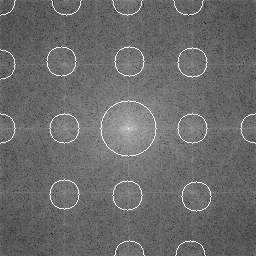}}
    %  \vspace{2.0cm}
      % \centerline{(b)}\medskip
    \end{minipage}

  \end{minipage}

  \caption{Fake generated paintings with noisy artifacts and its spectrum with the detected blobs.}
  \label{fig:blob}
\end{figure}

2) Artifacts: Inspired by \cite{DBLP:conf/mipr/MarraGVP19}, which discovered that certain noisy artifacts regularly appear in the GAN generated images, we analyze the GAN generated paintings in frequency domain. 
According to our experiments, grid-like artifacts, where each crossing point appears to be a blob, appear regularly in the spectrum of GAN based generated paintings. 
Fig.\,\ref{fig:blob} presents a generated painting with visible artifacts and its detected blobs.

\subsection{Statistical Feature}
\label{sec:stats}

Based on the first observation in Sec.\,\ref{sec:observation}, we
perform a simple yet effective univariate analysis to describe the statistical distribution of the spectrum.

Given a normalized spectrum denoted by $M$, let $S=(s_1,s_2,...,s_L)$ be the one-dimensional vector containing all 
the elements of $M$. Let $L$ denote the length of $S$. Four statistical values with different orders, including mean, standard deviation, skewness and kurtosis,
are calculated and then concatenated to form the statistical feature for FGPD-FA:
\begin{itemize}
  \item Mean: $\overline{s}=\frac{1}{L}\sum_{l=1}^{L}s_l$
  \item Standard deviation: $\sigma_s=\sqrt{\frac{1}{L-1}\sum_{l=1}^L (s_l - \bar{s})^2}$
  \item Skewness: $skewness = \frac{\sum_{l=1}^{L}(s_{l} - \bar{s})^{3}/L}{\sigma_s^{3}}$
  \item Kurtosis: $kurtosis = \frac{\sum_{l=1}^{L}(s_{l} - \bar{s})^{4}/L}{\sigma_s^{4}}  - 3$
\end{itemize}

\subsection{Oriented Gradient Feature}
\label{sec:hog}

According to Sec.\,\ref{sec:observation}, the changing slope of the spectrum of the fake generated paintings varies more obviously from low to high frequencies, compared to that of the real paintings. 
Thus, local oriented gradients can be exploited to model these inconsistencies.
FGPD-FA adopts the classic yet effective Histogram of Oriented Gradients (HOG) \cite {DBLP:conf/cvpr/DalalT05} to capture the local oriented gradients. 
Before the extraction of HOG vectors, considering the symmetry of the spectrum, FGPD-FA resizes the width and height of the 
spectrum by half and then crops the top-left quarter of the resized spectrum for further processing and feature extraction.
Then, a Gaussian filter is applied to smooth the resized and cropped spectrum. Although \cite {DBLP:conf/cvpr/DalalT05} shows that smoothing with large masks tends to decrease the performance, 
we empirically observe that it tends to be helpful to smooth the spectrum with large masks (e.g. $9\times 9$) in frequency domain. 
After the smoothing, 
we divide the preprocessed spectrum into cells. For each cell, a histogram of gradient directions is computed as a HOG descriptor.

To better represent the spectrum, with the computed HOG descriptors, we encode them into a Fisher vector \cite {DBLP:journals/ijcv/SanchezPMV13}.
For each image (spectrum), let $H=\{x_i|i=1,...,N\}$ be a set of D dimensional HOG features, where N denotes the number of cells and 
D denotes the number of orientations. Then, the corresponding Fisher vector can be computed by
\begin{equation}
  FV_{\Theta} = \frac{1}{N} \sum_{i=1}^N L_{\Theta} \frac{\partial \log{G(x_i | \Theta)}}{\partial \Theta},
\end{equation}
where $G$ denotes the probability distribution function of a Gaussian Mixture Model (GMM). Here, GMM is utilized 
to fit the distribution of the extracted HOG descriptors.
Note that $\Theta = \{w_k,\mu_k,\Sigma_k,k=1,...,K\}$ represents the parameters of GMM, where $K$ stands for the number of Gaussian distributions and $w_k,\mu_k,\Sigma_k$ represent 
the weight, mean vector and covariance matrix of Gaussian distribution k, respectively. $L_\Theta$ represents the Cholesky decomposition of the Fisher Information Matrix \cite{DBLP:conf/nips/JaakkolaH98}. 
In our experiments, a practical implementation of Fisher Vector Encoding \cite{fishervector} is employed for convenience.

\subsection{Blob Feature}
\label{sec:blob}

Since regular grid-like artifacts with blobs can be observed in frequency domain of the fake generated paintings, 
to better utilize this trace, we adopt the Difference of Gaussian (DOG) \cite {DBLP:journals/ijcv/Lowe04} method, 
which is an efficient approximation to the Laplacian of Gaussian (LOG) in blob detection \cite {DBLP:journals/ijcv/Lindeberg98}, 
to detect the blobs by finding the extrema. 
In our method, the number of detected blobs is exploited as the blob feature.

\subsection{Procedures of FGPD-FA}
Here, we briefly summarize the procedures of FGPD-FA, whose framework is shown in Fig.\,\ref{fig:framework}.
For both the training and testing stages, the given image is firstly transferred into Fourier frequency domain, where all the subsequent operations 
are performed. Next, the statistical feature, oriented gradient feature and blob feature are extracted and then concatenated into a 
feature vector. Support vector machine (SVM) is employed as the classifier. Note that the implementation of SVM in \cite{scikit-learn} is employed here. In the training process, GMM model is estimated based on the training data and 
SVM is trained with the extracted features and corresponding labels. In the testing process, the trained GMM and SVM are directly utilized in predictions.

\section{EXPERIMENTAL RESULTS}
\label{sec:expr}
In this section, our database, the experimental setups, evaluation results and an ablation 
study are correspondingly introduced in details.
\subsection{Database}
\label{sec:setup}
\begin{table}[]
  \caption{Details of the testing datasets.}
  \begin{center}
  \setlength{\tabcolsep}{1.5mm}{
    \begin{tabular}{|c|cccc|}
      \hline
      \textbf{Index} & \textbf{Artists} & \textbf{\begin{tabular}[c]{@{}c@{}}Forging \\ Algorithms\end{tabular}} & \textbf{\begin{tabular}[c]{@{}c@{}}Real \\ Images\end{tabular}} & \textbf{\begin{tabular}[c]{@{}c@{}}Fake \\ Images\end{tabular}} \\ \hline
      $T_\alpha ^ 1$            & Renoir           & All                 & 693                                                             & 708                                                             \\
      $T_\alpha ^ 2$             & Van Gogh         & All                 & 376                                                             & 705                                                             \\
      $T_\alpha ^ 3$            & Picasso          & All                 & 484                                                             & 705                                                             \\ \hline
      $T_\beta ^ 1$             & All              & AAMS \cite{DBLP:conf/cvpr/YaoRX0LW19}               & 1007                                                            & 1013                                                            \\
      $T_\beta ^ 2$             & All              & CycleGAN \cite{DBLP:conf/iccv/ZhuPIE17}            & 1007                                                            & 1044                                                            \\
      $T_\beta ^ 3$             & All              & GatedGAN \cite{DBLP:journals/tip/ChenXYST19}            & 1007                                                            & 1020                                                            \\ \hline
      $T_\gamma ^ 1$             & All              & All                 & 2006                                                            & 2092                                                            \\ \hline
      \end{tabular}
  }
\end{center}
  
  \label{table:data}
\end{table}

In this paper, we build a database, named Digital Painting Forgery Detection (DPFD) Database, which contains 58292 fake generated paintings and 10310 real paintings with the resolution of $256\times 256$.
The real paintings are from fifteen famous artists.
To generate the fake paintings, natural images are selected from multiple image datasets \cite {DBLP:conf/eccv/LinMBHPRDZ14,DBLP:conf/cvpr/PhilbinCISZ07,
DBLP:conf/cvpr/NilsbackZ06,DBLP:conf/cvpr/PhilbinCISZ08,DBLP:journals/ijcv/OlivaT01}.
The fake paintings are generated via three state-of-the-art style transfer methods, including 
AAMS \cite {DBLP:conf/cvpr/YaoRX0LW19}, CycleGAN \cite {DBLP:conf/iccv/ZhuPIE17}, and GatedGAN \cite {DBLP:journals/tip/ChenXYST19}.

In the experiments, both the training and testing datasets are constructed based on our DPFD database. 
Note that the training and testing datasets are not overlapping. 
The training dataset contains 4992 fake paintings with all the styles of fifteen artists, which are generated by all the three methods, and 4128 real paintings from all the fifteen artists. 
To evaluate the proposed method from different perspectives, seven testing datasets are constructed, as shown in Table\,\ref{table:data}.
Note that we only select three representative artists to assess the proposed method when it faces different forged styles.

\begin{table}[]
  \caption{Performance on different datasets.}
  \begin{center}
    \setlength{\tabcolsep}{0.5mm}{
      \begin{tabular}{|c|cclcccc|}
        \hline
        & \textbf{$T_\alpha^1$} & \textbf{$T_\alpha^2$} & \multicolumn{1}{c}{\textbf{$T_\alpha^3$}} & \textbf{$T_\beta^1$} & \textbf{$T_\beta^2$} & \textbf{$T_\beta^3$} & \textbf{$T_\gamma^1$}  \\ \hline
        \textbf{$Acc$(\%)} & 97.50       & 93.99       & 93.19                           & 95.05       & 94.10       & 97.68       & 94.85                               \\
        \textbf{$F_1$(\%)} & 97.48       & 91.91       & 91.91                           & 94.82       & 93.87       & 97.61      & 94.62                           \\ \hline
      \end{tabular}
    }
  \end{center}
  
  \label{table:perform}
\end{table}

\subsection{Setups and Metrics}
For the oriented gradient feature, we set the Gaussian filter size as $9 \times 9$, the number of orientations as 9, the cell size as $16 \times 16$, the block size and the window size as 
$64 \times 64$. L2-norm is employed as the block normalization method. Note that the power law equalization is not applied. Besides, when fitting GMM, the number of Gaussian distributions are chosen 
 to be 16 by Bayesian Information Criterion \cite {schwarz1978estimating}. For the blob feature, to neglect the very small blobs, we set the minimum standard deviation for the Gaussian kernel as 11, 
the maximum standard deviation as 30, and the ratio as 1.8. In SVM, the RBF kernel is selected.
 The regularization parameter and the kernel coefficient are selected by grid search strategy.

In the evaluation, the accuracy ($Acc$) and F1-score ($F_1$) are employed as measurements \cite{hossin2015review}.
Note that the fake generated paintings are defined as negative samples and the real paintings are defined as positive samples.

\subsection{Quantitative Evaluation Results}

The proposed FGPD-FA is evaluated with the datasets in Table\,\ref{table:data} and the quantitative results are shown in Table\,\ref{table:perform}.
The results indicate that our method performs decently for both the CNN-based method ($T_\beta^1$) and GAN-based methods ($T_\beta^2$, $T_\beta^3$),
and it can be used to detect the fake generated paintings with specific styles ($T_\alpha^1$-$T_\alpha^3$).
When the fake images are generated based on all the fifteen artists and three forging methods, FPGD-FA still gives satisfactory performance, 
which demonstrates the excellence of the proposed FPGD-FA.

\subsection{Ablation Study}

\begin{figure}
  \centering
  \begin{minipage}[b]{0.79\linewidth}
    \centering
    \centerline{\includegraphics[width=\linewidth]{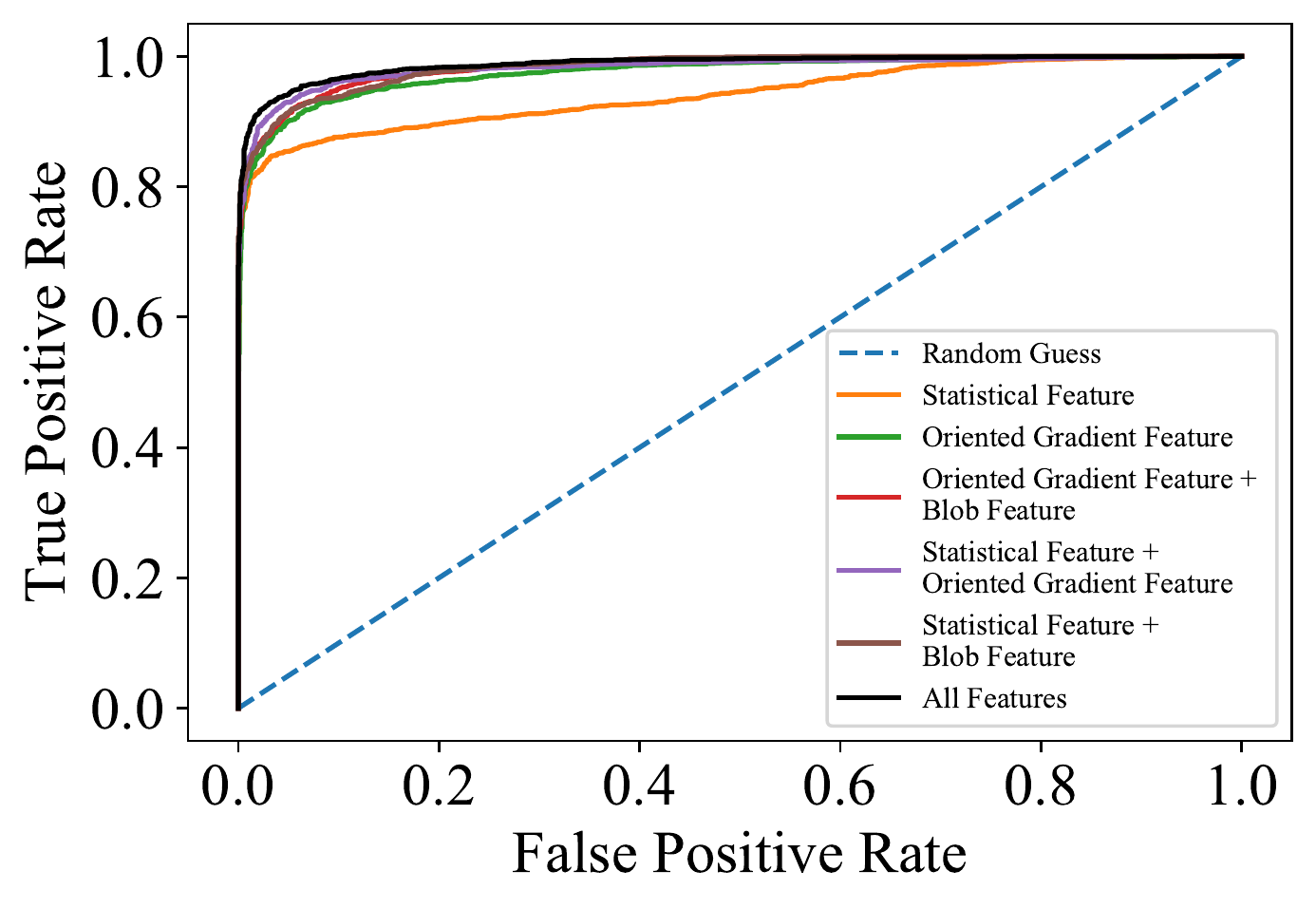}}
  %  \vspace{2.0cm}
    % \centerline{(a)}\medskip
  \end{minipage}
  \caption{ROC curves for dataset $T_\gamma^1$.}
  \label{fig:roc}
  \end{figure}

To verify the effectiveness of each feature, an ablation study is performed on the testing dataset $T_\gamma^1$. 
The results are reported in Table\,\ref{table:ablation} and the corresponding ROC curves are shown in Fig.\,\ref{fig:roc}.
As can be observed, the proposed method can identify the fake generated paintings from the real paintings with either the statistical feature or the oriented gradient feature.
Note that the blob feature is not evaluated alone because it is mainly designed to detect the GAN-based generated paintings.
When each two of the features jointly work together, obvious improvements can be obtained.
When all the three features are utilized, 
the proposed FGPD-FA gives the best performance. According to the results, the effectiveness of each feature can be verified.

\begin{table}[]
  \caption{Ablation Study.}
  \begin{center}
    \setlength{\tabcolsep}{0.6mm}{
      \begin{tabular}{|c|cccccc|}
        \hline
        \textbf{\begin{tabular}[c]{@{}c@{}}Statistical \\ Feature\end{tabular}}       & \checkmark &                           &                           & \checkmark & \checkmark & \checkmark \\ \hline
        \textbf{\begin{tabular}[c]{@{}c@{}}Oriented Gradient \\ Feature\end{tabular}} &                           & \checkmark & \checkmark & \checkmark &                           & \checkmark \\ \hline
        \textbf{\begin{tabular}[c]{@{}c@{}}Blob \\ Feature\end{tabular}}              &                           &                           & \checkmark &                           & \checkmark & \checkmark \\ \hline
          \hline
        \textbf{Acc(\%)}                                                               & 89.87                     & 92.31                     & \multicolumn{1}{c}{92.90} & 93.94                     & 92.65                     & \textbf{94.85}            \\
        \textbf{F1(\%)}                                                                & 88.60                     & 91.87                     & \multicolumn{1}{c}{92.53} & 93.64                     & 92.15                     & \textbf{94.62}            \\ \hline
        \end{tabular}
    }
  \end{center}
  \label{table:ablation}
 \end{table}

\section{CONCLUSION}
\label{sec:print}

In this paper, we aim to design a forgery detection approach to identify the fake generated paintings, which are forged by style transfer techniques, 
from the real paintings. We firstly analyze the fake generated and real paintings in Fourier frequency domain. Based on two major 
observations, we propose Fake Generated Painting Detection via Frequency Analysis by extracting the statistical, oriented gradient
 and blob features in frequency domain. We also propose a new database for fake generated painting detection. Extensive experiments 
 not only demonstrate the excellence of the proposed FGPD-FA, but also verify the effectiveness of each extracted feature.
\newpage

\bibliographystyle{IEEEbib}
\bibliography{refs}

\end{document}